\documentclass{article}
\usepackage{ijcai22}

\usepackage{times}

\usepackage{soul}
\usepackage{url}
\usepackage[hidelinks]{hyperref}
\usepackage[utf8]{inputenc}
\usepackage[small]{caption}
\usepackage[pdftex]{graphicx}
\usepackage{amsmath}
\usepackage{booktabs}
\usepackage{xcolor}
\usepackage{amsfonts}
\usepackage{epsfig}
\usepackage{mathtools}
\usepackage{multirow}
\usepackage{algorithm}
\usepackage{algorithmic,eqparbox,array}
\usepackage{booktabs}
\usepackage{makecell}
\newcommand{\etal}{\textit{et al}.}
\title{Iterative Geometry-Aware Cross Guidance Network for Stereo Image Inpainting}

\author{
Ang Li$^1$\and
Shanshan Zhao$^2$\and
Zhang Qingjie$^1$\And
Qiuhong Ke$^{3}$\footnote{Corresponding author.}\\
\affiliations
$^1$Aviation University of Air Force\\
$^2$JD Explore Academy, JD.COM\\
$^3$The University of Melbourne\\
\emails
angli.cs@outlook.com,
sshan.zhao00@gmail.com,
nudtzhang@hotmail.com,
qiuhong.ke@unimelb.edu.au
}

\begin{document}

\maketitle

\begin{abstract}

Currently, single image inpainting has achieved promising results based on deep convolutional neural networks. However, inpainting on stereo images with missing regions has not been explored thoroughly, which is also a significant but different problem. One crucial requirement for stereo image inpainting is stereo consistency. To achieve it, we propose an Iterative Geometry-Aware Cross Guidance Network (IGGNet).
The IGGNet contains two key ingredients, i.e., a  Geometry-Aware Attention (GAA) module and an Iterative Cross Guidance (ICG) strategy. The GAA module relies on the epipolar geometry cues and learns the geometry-aware guidance from one view to another, which is beneficial to make the corresponding regions in two views consistent. However, learning guidance from co-existing missing regions is challenging. To address this issue, the ICG strategy is proposed, which can  alternately narrow down the missing regions of the two views in an iterative manner.
Experimental results demonstrate that our proposed network outperforms the latest stereo image inpainting model and state-of-the-art single image inpainting models.
\end{abstract}

\section{Introduction}\label{secIntro}
Image inpainting aims to synthesize alternative contents for the missing regions of an image while simultaneously preserving the overall structure.
It has a wide range of applications such as photo recovery and object removal.
In recent years, relying on the powerful representation capabilities, deep learning based methods~\cite{liu2018image,Yu_2019_ICCV,Li_2020_CVPR,Zeng_2021_ICCV} have been dominating for image inpainting and are able to generate promising results. 
Apart from single images, stereo images are also important media due to the use of head-mounted devices ({\it e.g.,} AR/VR glasses), dual-lens smart phones, and stereo cameras. As a result, in some scenarios such as object removal and content editing, we are also required to recover the missing regions in the stereo images. However, inpainting stereo images has not been explored thoroughly yet. 

\begin{figure}[t!]
    \centering
    \setlength{\tabcolsep}{0.05em}
    {
\begin{tabular}{ccccc}
        
\rotatebox[origin=l]{90}{\small Left}&
\includegraphics[width=1.9cm]{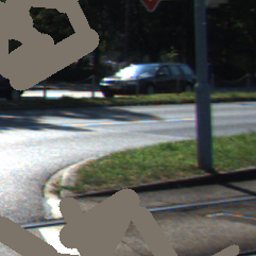}&
\includegraphics[width=1.9cm]{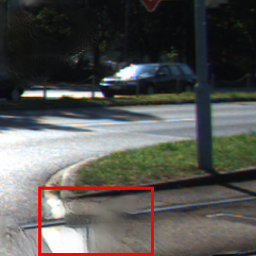}&
\includegraphics[width=1.9cm]{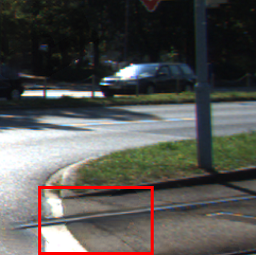}&
\includegraphics[width=1.9cm]{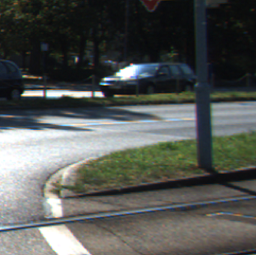}\\

\addlinespace[-0.20em]

\rotatebox[origin=l]{90}{\small Right}&
\includegraphics[width=1.9cm]{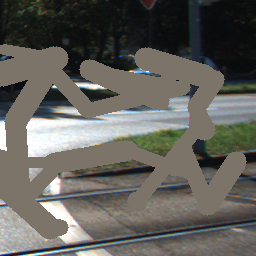}&
\includegraphics[width=1.9cm]{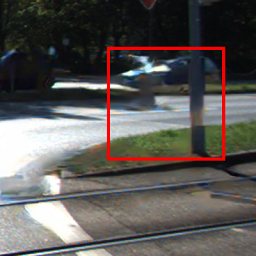}&
\includegraphics[width=1.9cm]{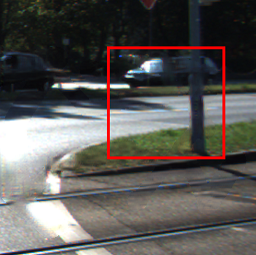}&
\includegraphics[width=1.9cm]{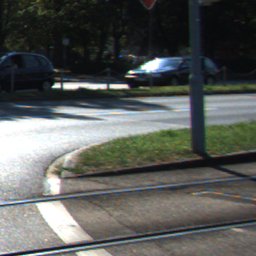}\\

  & \small Input   & \small SICNet    & \small IGGNet   & \small  GT \\
  
\end{tabular}  

}
\caption{An example of visual comparison with state-of-the-art stereo image inpainting model SICNet \protect\cite{Ma2020LearningAV} on a stereo pair (which contains two images, \protect\textit{i.e.,} left view and right view) from the  KITTI2015 dataset \protect\cite{mayer2016large}. GT is the ground truth image. As shown here, the input stereo pair can have different missing regions in each view. The proposed IGGNet is effective to incorporate geometry correlation and generate stereo consistent results.}
\label{example}
\end{figure}

A straightforward solution of stereo image inpainting is to directly apply single image inpainting methods (developed for individual images) over the left view and the right view separately. 
However, this may lead to severe stereo inconsistency, since inpainting an individual image only considers the undamaged spatial statistics in that image, {\it i.e.,} only one view of the stereo images, but ignores the geometric relationship between the two views. 
Here, stereo inconsistency refers to that the inpainted stereo pair, {\it i.e.,} the left and the right output images, has different structures or textures in the corresponding region. To alleviate this issue, Ma \etal~\shortcite{Ma2020LearningAV} recently propose SICNet for stereo image inpainting, which associates the two views via a feature map concatenation operation. Although simple, the concatenation cannot fully explore the stereo cues, and thus the stereo inconsistency is not addressed well. We give an example in Figure~\ref{example}.

To achieve better stereo consistency, we consider the epipolar geometry existing in the stereo pair. In particular, the epipolar geometry refers to that most of contents in one view can be found in another view. Therefore, this cue can motivate us to associate the features of potential corresponding regions in the stereo pair. In this way, if a region is missing in one view but exists in another, this cue can help us recover the region through querying related contents from the other view. Additionally, this cue is also helpful to make the structure and texture of corresponding regions consistent. 
However, for the co-existing region, \textit{i.e.,} a region missing in both views, it is still challenging to recover the content using the epipolar-geometry-based feature association
since it is difficult to model the geometric correspondence under this situation.

Motivated by the above observations and analysis, we propose a stereo inpainting network named \textbf{Iterative Geometry-Aware Cross Guidance Network (IGGNet)}. 
This network performs inpainting on the stereo images through exploring and integrating the stereo geometry in an iterative manner. IGGNet consists of two key ingredients, {\it i.e.,}   the \emph{Geometry-Aware Attention} (GAA) module and the \emph{Iterative Cross Guidance} (ICG) strategy.  
Relying on the epipolar geometry cues, the GAA module associates the two views through explicitly learning attention maps.
In such manner, it is capable of adaptively capturing the useful information from the reference view to guide the restoration of the target view without the requirement of disparity information. 
The ICG strategy is implemented to alternately narrow down the missing regions in the two views based on a confidence-driven mask updating policy. 
It is beneficial to alleviate the impact of co-existing missing regions and thus obtain better visual consistency in the inpainted results.
More specifically, in a certain iteration, we take one view as the target view (the view to be inpainted) and the other one as the reference (guidance). 
Then, given the inpainted target view obtained in this iteration and the reference view, 
we swap their roles (reference and target), apply the updated masks to them, and repeat such cross-guidance process for the following iterations. 
By processing the two views using the geometry-aware cross guidance in an iterative manner, our model is able to yield high-quality stereo image inpainting results with better geometry consistency.

In summary, our main contributions are:
\begin{itemize}

  \item We propose a novel end-to-end network for stereo image inpainting, which models the association between the two views through learning geometry-aware guidance and crossly performs inpainting in an iterative manner.

  \item We introduce a \textit{Geometry-Aware Attention} (GAA) module that incorporates the correlation between the two views of a stereo pair based on the epipolar geometry. 
  The GAA module relies on the epipolar geometry cue and benefits generating stereo consistent results.
  
  \item We adopt an \textit{Iterative Cross Guidance} (ICG) strategy that  exploits the geometry-aware guidance to alternately shrink the missing regions of the two views in an iterative manner. In this way, we can deal with co-existing missing regions better in the stereo pair.

  \item Experiments on real data demonstrate that our proposed IGGNet outperforms state-of-the-art single image inpainting methods and  stereo image inpainting methods.
\end{itemize}

\section{Related Work}
\subsection{Image Inpainting}
Different from conventional image inpainting approaches, deep learning-based image inpainting models can generate more visually plausible details or fill large missing regions with new contents that never exist in the input image \cite{pathak2016context,iizuka2017globally,yu2018generative,liu2018image,Ang_2019_IJCNN,Yu_2019_ICCV,Liu_2019_ICCV,li2019generative,Li_2020_CVPR,Yi_2020_CVPR,Zeng_2021_ICCV}.
An advantage of these models is to adaptively synthesize visually plausible contents for different semantics.
Pathak \etal~\shortcite{pathak2016context} first introduce the \textit{Context Encoder} (CE) model where a convolutional encoder-decoder network is trained with the combination of an adversarial loss \cite{goodfellow2014generative} and a reconstruction loss.
Yu \etal~\shortcite{Yu_2019_ICCV} present DeepFillv2 to provide a learnable dynamic mask-updating mechanism for free-form image inpainting.
Zeng \etal~\shortcite{Zeng_2021_ICCV} propose CR-Fill with a learnable contextual reconstruction loss.
Since these single image inpainting models only take one view as input and ignore the epipolar geometry correlation when dealing with stereo images, they are unable to generate stereo consistent results.

\subsection{Stereo Image Inpainting}
{
Several conventional methods~\cite{wang2008stereoscopic,hervieu2010stereoscopic,morse2012patchmatch} have been proposed to address stereo image inpainting.
However, these methods fail to generate meaningful structures when facing complex semantic scenes in the missing regions, which is also a shared limitation for conventional single image and video inpainting methods.
Chen \etal~\shortcite{Chen2019CNNBasedSI} design the first CNN-based model for stereo image inpainting based on ~\cite{pathak2016context}. 
However, this method can only deal with square holes in the centre. 
Ma \etal~\shortcite{Ma2020LearningAV} propose the SICNet with a cross-fusion design.
However, the fusion module simply concatenates the two views' feature maps and it is unable to effectively explore the stereo cues.
}

\section{Proposed Method}
In this section, we first present the overall network architecture.
Then, we describe the GAA module and  the ICG strategy in more details, respectively.
Finally, the training target functions are introduced.

\begin{figure*}[tb]
\centering
\includegraphics[width=0.8\linewidth]{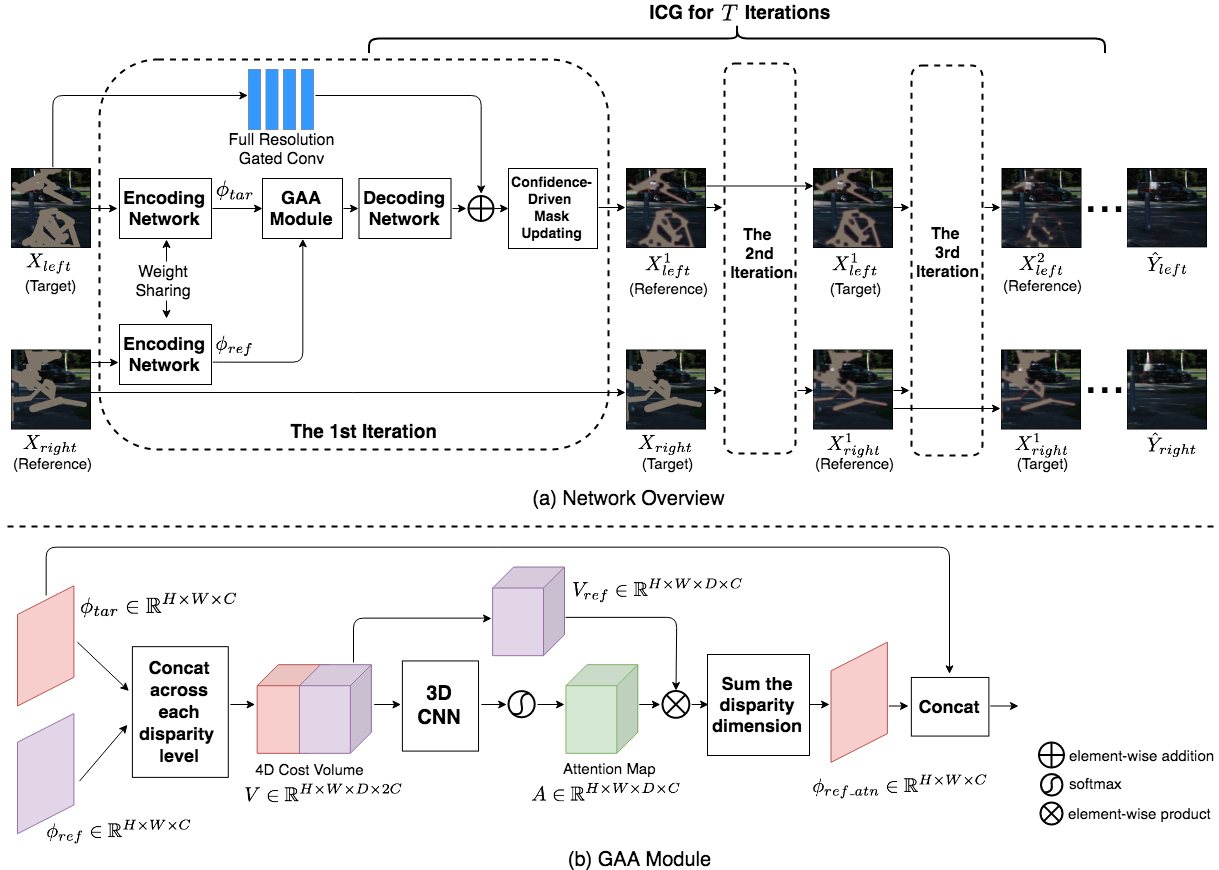}
\caption{Our proposed IGGNet network contains two key parts, a Geometry-aware Attention (GAA) module and an Iterative Cross Guidance (ICG) strategy. The
Geometry-aware Attention (GAA) module explores the geometry correlation of the stereo pair and performs attention-driven feature aggregation. The Iterative Cross Guidance (ICG) strategy exploits the guidance through a confidence-driven mask updating module to alternately shrink the missing regions of the two views.}
\label{fig:structure}

\end{figure*}

\subsection{Network Overview}
Given a rectified stereo image pair $(X_{left}, X_{right})$ annotated with binary masks $(M_{left}, M_{right})$ which indicate the missing region, the goal of our network is to generate the output pair $(\hat{Y}_{left},\hat{Y}_{right})$.
The output pair should not only have visually plausible restored contents but also maintain the geometry consistency with the ground truth pair $(Y_{left},Y_{right})$.
To achieve this goal, we restore the two images in an iterative manner with the proposed \textbf{Iterative Cross Guidance (ICG)} strategy  using the deep network containing a \textbf{Geometry-Aware Attention (GAA)} module.

Specifically, 
for the first iteration, we take the left view $X_{left}$ as the target (the view to be inpainted) and the right view $X_{right}$ as the reference (guidance).
As shown in Figure~\ref{fig:structure} (a), the IGGNet includes two  parallel branches, \textit{i.e.,} an encoder-decoder branch and a full resolution branch.
We apply Gated convolution \cite{Yu_2019_ICCV} as the basic convolution operation in both branches.
For the encoder-decoder branch, both the target and reference images are fed into the encoding network, which generates the corresponding feature maps $\phi_{tar}$ and $\phi_{ref}$.
The two feature maps are then processed by the  
GAA module, which explores the geometry correspondence between them and incorporates most relevant context from $\phi_{ref}$ to guide the inpainting of $\phi_{tar}$ (More details are explained in Section~\ref{secGAA}).  
Then, the decoding network decodes the aggregated feature maps (output of the GAA module) to a restored target image. In addition,
the full resolution branch consists of four  consecutive full-resolution Gated convolution \cite{Yu_2019_ICCV} layers, which are beneficial to the maintenance of high-resolution representations \cite{Guo2019ProgressiveII}.
The outputs of two parallel branches are then fused through the element-wise addition operation. 

After the first iteration, we obtain the inpainted left image $X_{left}^{1}$ with narrowed missing regions, as shown in Figure~\ref{fig:structure} (a).
Next, we swap the roles of the two images, {\it i.e.,} the inpainted left image $X_{left}^{1}$ as the reference and the original right images $X_{right}$ as the target, and then feed them to the second iteration, which yields the inpainted right image $X_{right}^1$.  
In the following iterations, we continue to swap the roles of target and reference, {\it e.g.,} $X_{left}^1$ (target) and $X_{right}^1$ (reference) as input for the third iteration, and repeat the cross guidance process over $T$ iterations in total to obtain the final restored images $\hat{Y}_{left}$ and $\hat{Y}_{right}$. In the following, we detail the two main involved components.


\subsection{Geometry-Aware Attention Module}\label{secGAA}
In this part, we introduce the GAA module by taking the example of the left image as the target and the right as the reference. Since the missing content of the left image might occur in the right one, it is crucial to exploit the guidance from the right image to inpaint the left. To achieve this, we can warp the right image according to the disparity map~\cite{liang2018learning}, which indicates the displacement between the corresponding pixels of the two views, and then directly exploit the warped image as the guidance. However, the disparity information is often unavailable, which can be obtained by traditional SGM algorithms~\cite{hirschmuller2005accurate} or deep learning based approaches~\cite{liang2018learning}. Both of them require extra computational and time costs. 
Additionally, if there are  missing regions or noises in the disparity map, we also need to process the disparity map firstly. 
To dispense with this needless step, we examine the epipolar geometry existing in the stereo images. In specific, for a rectified stereo image pair, there only exists the horizontal displacement between the corresponded pixels in the two views. As a result, for a pixel in one view, we can model the relationship between it and its potential correspondences through learning an attention map upon the $4$D cost-volume created from the feature maps of the two views.
As a result, our model is able to adaptively capture the useful information from the reference to guide the restoration of the target without the requirement of disparity information. 
Next, we provide more details.

As shown in Figure~\ref{fig:structure} (b), the GAA module first builds a cost volume upon the encoded feature maps $\phi_{tar}\in \mathbb{R}^{H\times W \times C}$ and $\phi_{ref} \in \mathbb{R}^{H\times W \times C}$.
Specifically, $\phi_{ref}$ is shifted by different disparity levels ($\{0,1,…,D-1\}$) along the horizontal direction and then concatenated with $\phi_{tar}$, which is repeated $D$ times to generate a 4D volume $V \in \mathbb{R}^{H\times W \times D \times 2C}$.
Here $D$ denotes the predefined maximum disparity value.
In order to aggregate the feature information 
along the disparity dimension as well as the channel dimension, 
we first apply consecutive 3D convolutional operations on the channel dimension followed by a Softmax operation on the disparity dimension to get a $4$D attention map $A \in \mathbb{R}^{H\times W \times D \times C}$. 
We then perform an attention-driven feature aggregation to update the features of the target adaptively.
Specifically, we apply element-wise product on $A$ and $V_{ref}\in\mathbb{R}^{H\times W \times D \times C}$, where $V_{ref}$ represents the corresponding part of the reference view in $V$,
and sum the elements  along the disparity dimension (the third dimension) to obtain the attended reference feature map $\phi_{ref\_atn} \in \mathbb{R}^{H\times W \times C}$:

\begin{equation}
\phi_{ref\_atn} = \textrm{sum}(A\cdot V_{ref}, dim=3),
\end{equation}
where $\cdot$ denotes the element-wise product.

Finally, we concatenate $\phi_{tar}$ and $\phi_{ref\_atn}$ on the channel dimension as the output of GAA module.

\subsection{Iterative Cross Guidance}
The co-existing missing regions in the stereo image pair make the inpainting task extremely challenging. When the model uses the reference view as guidance to inpaint the target view, missing regions in the reference view inevitably causes inaccurate attention computation and feature aggregation.
Therefore, we propose an iterative cross guidance inpainting strategy. 
The whole inpainting procedure is composed of a number $T$ of inpainting iterations, in which the two views alternately serve as the target view and reference view, respectively.
Each view is inpainted by $T/2$ times ($T$ is supposed to be an even number).
In each iteration, we shrink the missing region in the target view with a confidence-driven mask updating module.
Note that we use the same IGGNet in all the iterations.

\noindent\textbf{Confidence-Driven Mask Updating} 
We observe that in the failure cases of existing approaches, despite the artifacts,
there often exist sub-regions with plausible predictions. 
By trusting the plausible part and treating the remaining part as a new mask, we can run the network again and again.
In this way, the mask becomes progressively smaller and the model can produce better results.
Inspired by this observation, we exploit a confidence-driven mask updating module to keep the inpainted region with higher confidence in the target image but ignore those with large prediction error.
To realize the mask confidence prediction, we utilize the Gated Convolution \cite{Yu_2019_ICCV} in both the full-resolution and the encoder-decoder branches instead of the standard 2D convolution.
Specifically, we modify the last Gated convolution layer of the two branches by setting a confidence threshold on the generated three-channel soft mask $S \in [0,1]^{H\times W \times 3}$, and obtain the updated one-channel binary mask $M\in \{0,1\}^{H\times W \times 1}$.

\begin{equation}
  M(i,j)=\left\{
  \begin{array}{@{}ll@{}}
    1, & \text{if}\ \max\limits_{d=0,1,2}(S(i,j,d))>0.5 \\  
    0, & \text{otherwise}
  \end{array}\right.
 \label{eq:mask}
\end{equation} 
We represent this process as a function $g$, \textit{i.e.,} $M=g(S)$.

Algorithm 1 illustrates the procedure of the ICG strategy.

\begin{algorithm}[H]
\algsetup{linenosize=\small}
\caption{Iterative Cross Guidance}
\label{ALG1}
\small
\begin{algorithmic}[1]
\REQUIRE number of iterations $T$, left input image $X_{left}$, right input image $X_{right}$, full resolution branch $\text{Branch\_f}$, encoder-decoder branch $\text{Branch\_ed}$ 
\ENSURE restored left image $\hat{Y}_{left}$, restored right image $\hat{Y}_{right}$ 
\STATE $X_{tar}^{0} = X_{left}, X_{ref}^{0} = X_{right}$
\FOR{$t$ in ${1,...,T}$}
\STATE $ R_{f}, S_{f} = \text{Branch\_f}(X_{tar}^{t-1})$
\STATE $ R_{ed}, S_{ed} = \text{Branch\_ed}(X_{tar}^{t-1}, X_{ref}^{t-1})$
\STATE $M_{f} = g(S_{f})\ (\textrm{Eq}.~\ref{eq:mask})$
\STATE $M_{ed} = g(S_{ed})\ (\textrm{Eq}.~\ref{eq:mask})$
\STATE $M_{t} = M_{f} \cdot M_{ed}$
\STATE $R^{t} = \frac{1}{2}(M_{t} \cdot (R_{f} + R_{ed}))$
\STATE $X_{tar}^{t} = X_{ref}^{t-1}$
\STATE $X_{ref}^{t} = R^{t}$
\IF{$t \%  2 == 1$}
\STATE $X_{left}^{(t+1)/2} = R^{t}$
\ELSE
\STATE $X_{right}^{t/2} = R^{t}$
\ENDIF
\ENDFOR
\STATE $\hat{Y}_{left} = X_{left}^{T/2}, \hat{Y}_{right} = X_{right}^{T/2}$

\end{algorithmic}
\end{algorithm}

\subsection{Loss Function}
We train our IGGNet following a typical GAN~\cite{goodfellow2014generative} procedure. The loss function used for training is:

\begin{equation}
\mathcal{L}_{total} = \mathcal{L}_{\text{rec}} + \lambda_{\text{adv}}\mathcal{L}_{\text{adv}}.
\end{equation}
Here, $\mathcal{L}_{\text{rec}}$ and $\mathcal{L}_{\text{adv}}$ denote reconstruction loss and adversarial loss respectively.
The balancing weight $\lambda_{\text{adv}}$ is empirically set to 0.01. The formulation of  $\mathcal{L}_{\text{rec}}$ and $\mathcal{L}_{\text{adv}}$ are as follows:

\begin{equation}
\mathcal{L}_{\text{rec}} = \sum_{t}^{T/2}\|(X_{left}^{t} - Y_{left}) + (X_{right}^{t} - Y_{right})\|_{1},
\end{equation}

\begin{equation}
\begin{split}
\mathcal{L}_{\text{adv}} & = \sum_{t}^{T/2} (\log(Dis(Y_{left})) + \log(1 - Dis(X_{left}^{t})) \\ & + \log(Dis(Y_{right})) + \log(1 - Dis(X_{right}^{t})) ),
\end{split}
\end{equation}
where $Dis$ denotes the discriminator of GAN. We also adopt spectral normalization \cite{Miyato2018SpectralNF} to stabilize the convergence of the discriminator.
Here the generator of the GAN is IGGNet.
Details of network structures are provided in the supplementary material.
\section{Experiments}
We evaluate our method and make comparisons against state-of-the-art single image and stereo image inpainting models qualitatively and quantitatively. 
A comprehensive ablation study is also conducted on our proposed model.

\begin{table}[thb]
\renewcommand\arraystretch{1.05}

\begin{center}
\scalebox{0.59}
{
\begin{tabular}{l|ccc|ccc|ccc}
      			
      			\hline
      			\multicolumn{10}{c}{KITTI2015} \\
      			\hline
      	 \multirow{2}{*}{Model}  & \multicolumn{3}{c|}{mask 0-20\%} & \multicolumn{3}{c|}{mask 20-40\%} & \multicolumn{3}{c}{mask 40-60\%}  \\ 
     			& PSNR & SSIM & FID  & PSNR & SSIM & FID & PSNR & SSIM & FID \\ \hline 
     	DeepFillv2 & 27.15 & 0.919 & 6.42 & 21.91 & 0.844 & 12.37 & 17.68 &  0.694 & 26.56  \\  
		CR-Fill  & 27.84 & 0.927 & 5.86 & 22.45  & 0.863 & 11.60 & 17.89 & 0.711 & 25.18 \\	 
		SICNet	 & 28.18 & 0.935 & 5.47 & 22.74  & 0.852 & 10.48 & 18.04 & 0.706 & 23.65 \\ \hline
		\textbf{IGGNet (w/o ICG)} & 28.45 & 0.932 & 5.18 &  23.37  & 0.860 & 9.91 & 18.81 & 0.714 & 22.47  \\ 
      		\textbf{IGGNet} & \textbf{29.31} & \textbf{0.941} & \textbf{4.69} &  \textbf{24.20}  & \textbf{0.881} & \textbf{8.07} & \textbf{19.58} & \textbf{0.723} & \textbf{20.24}  \\ 
      		\hline
      		\hline
      			\multicolumn{10}{c}{MPI-Sintel} \\
      			\hline
      	 \multirow{2}{*}{Model}  & \multicolumn{3}{c|}{mask 0-20\%} & \multicolumn{3}{c|}{mask 20-40\%} & \multicolumn{3}{c}{mask 40-60\%}  \\ 
     			& PSNR & SSIM & FID  & PSNR & SSIM & FID & PSNR & SSIM & FID \\ \hline 
     	DeepFillv2	 & 28.40 & 0.918 & 5.51 & 23.26 & 0.838 & 13.64 & 18.11 &  0.730 & 25.54  \\
		CR-Fill  & 29.07 & 0.935 & 4.61 & 24.03  & 0.884 & 10.47 & 18.29 & 0.746 & 23.72  \\	 
  
		SICNet 	 & 29.87 & 0.944 & 3.97 & 24.21  & 0.876 & 8.83 & 18.65 & 0.743 & 22.40  \\ \hline
		\textbf{IGGNet (w/o ICG)} & 30.25 & 0.947 & 4.08 &  24.72  & 0.885 & 8.47 & 19.50 & 0.754 & 21.64   \\ 
      		\textbf{IGGNet} & \textbf{31.02} & \textbf{0.951} & \textbf{3.75} &  \textbf{25.56}  & \textbf{0.903} & \textbf{7.29} & \textbf{20.81} & \textbf{0.768} & \textbf{19.82}   \\ 
      		\hline
\end{tabular}    
}
\end{center}
\caption{Quantitative comparisons on KITTI2015 and MPI-Sintel datasets under three different mask ratio settings regarding 3 metrics: PSNR (higher is better), SSIM (higher is better) and FID (lower is better). Here, ``w/o'' means ``without''. The best results are in \textbf{bold}.}
\label{tab0}
\end{table}

\noindent\textbf{Datasets.}
We train and evaluate our model on two well-known stereo image datasets: \textbf{KITTI2015} \cite{mayer2016large} and \textbf{MPI-Sintel} \cite{Butler_ECCV_2012}.
For the KITTI2015 dataset, we use the original training and testing splits.
For the MPI-Sintel dataset, we select 16 video directories for training, and the other 7 for testing.
All training images are randomly cropped to $256 \times 256$ for every epoch.
Each testing image in the KITTI2015 dataset is horizontally split to 4 images and resize them to $256 \times 256$, while that in the MPI-Sintel dataset is directly resized to $256 \times 256$.
It is worth noting that we do not apply any pre-processing or post-processing operation. 
In our experiments, we define the maximum image disparity to be 192. 
Since we apply GAA module on the $1/4$ scale level, $D$ is set as $48$.

\noindent\textbf{Mask Settings.}
{To simulate diverse scenarios, we use the irregular mask dataset \cite{Yu_2019_ICCV} that contains masks with arbitrary shapes and random locations.
}
For each stereo image pair, we apply different masks on the two views, respectively.
We use three different mask ratios (mask area relative to the entire image): $0 - 20\%$, $20 - 40\%$ and $40 - 60\%$.

\noindent\textbf{Baseline Models.}
We compare our model with two state-of-the-art single image inpainting models \textbf{DeepFillv2}~\cite{Yu_2019_ICCV} and \textbf{CR-Fill} \cite{Zeng_2021_ICCV}, and the latest stereo image inpainting model \textbf{SICNet} \cite{Ma2020LearningAV}.
For fair comparison, we train and test the baseline models and our model with the same dataset (KITTI2015 or MPI-Sintel). For DeeFillv2 and CR-Fill, we use their released code but not use their open-source pretrained weights, For SICNet, we reproduce it according to the descriptions in its paper.

\subsection{Quantitative Results}
We consider three metrics for evaluation: 1) \textit{Peak Signal-to-Noise Ratio} (PSNR, measures image distortion), 2) \textit{Structural Similarity} (SSIM, measures structure similarity) and 3) the \textit{Fréchet Inception Distance} (FID, measures perceptual quality) \cite{heusel2017gans}. 
As shown in Table~\ref{tab0}, even without ICG strategy, our model still achieves superior results over the baseline models, which indicates that GAA can effectively aggregate stereo correspondence into the inpainting process.
When utilizing ICG strategy, our model has a more obvious lead, especially when dealing with larger masks (which contain more co-existing missing regions).
This verifies that ICG can alleviate the problem of co-existing missing regions.

\subsection{Qualitative Results}
\begin{figure*}[t!]

    \centering
    \setlength{\tabcolsep}{0.05em}
    {

\begin{tabular}{ccccccccccccc}

\rotatebox[origin=l]{90}{\scriptsize GT}&
\includegraphics[width=1.4cm]{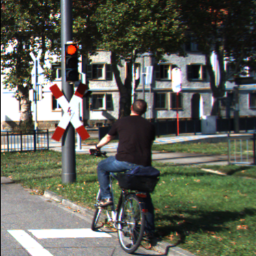}&
\includegraphics[width=1.4cm]{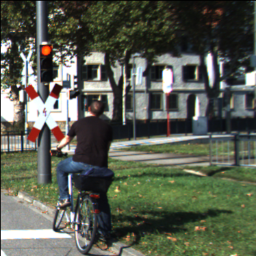}&
\includegraphics[width=1.4cm]{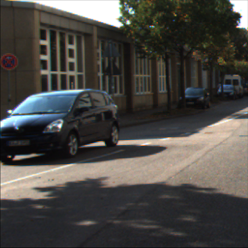}&
\includegraphics[width=1.4cm]{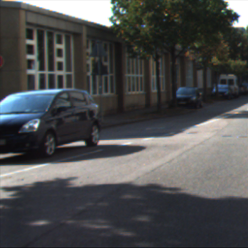}&
\includegraphics[width=1.4cm]{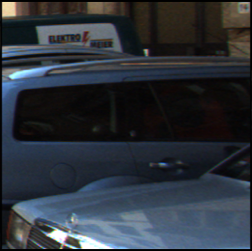}&
\includegraphics[width=1.4cm]{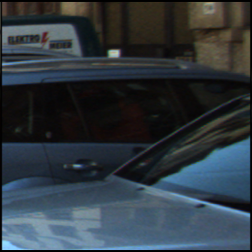}&
\includegraphics[width=1.4cm]{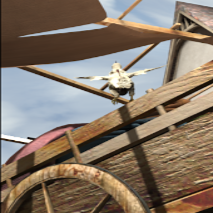}&
\includegraphics[width=1.4cm]{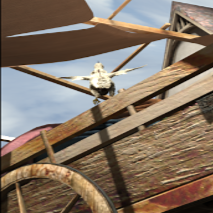}&
\includegraphics[width=1.4cm]{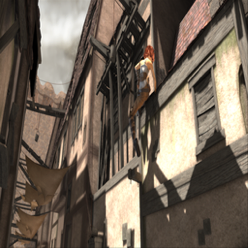}&
\includegraphics[width=1.4cm]{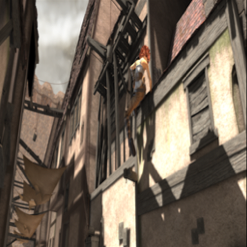}&
\includegraphics[width=1.4cm]{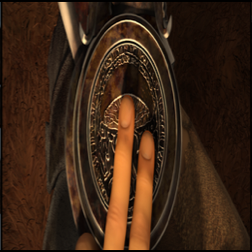}&
\includegraphics[width=1.4cm]{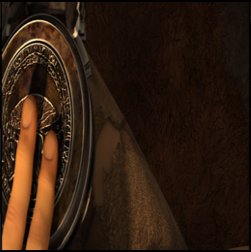}
\\

\addlinespace[-0.25em]
        
\rotatebox[origin=l]{90}{\scriptsize Input}&
\includegraphics[width=1.4cm]{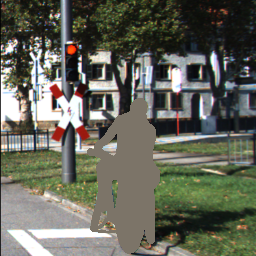}&
\includegraphics[width=1.4cm]{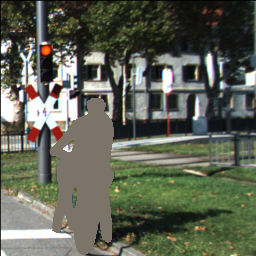}&
\includegraphics[width=1.4cm]{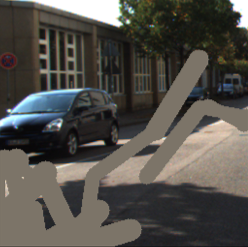}&
\includegraphics[width=1.4cm]{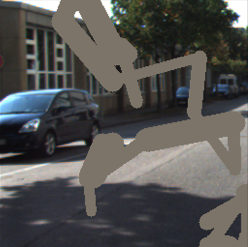}&
\includegraphics[width=1.4cm]{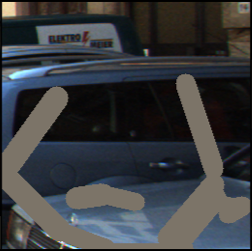}&
\includegraphics[width=1.4cm]{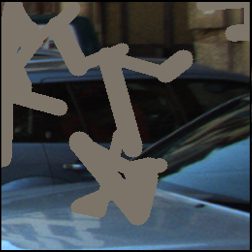}&
\includegraphics[width=1.4cm]{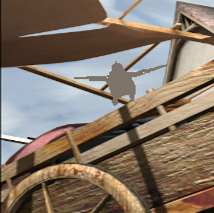}&
\includegraphics[width=1.4cm]{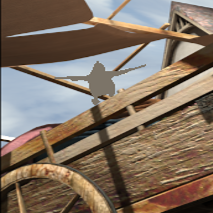}&
\includegraphics[width=1.4cm]{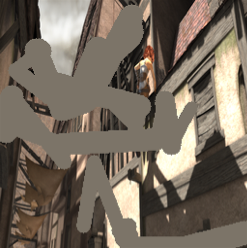}&
\includegraphics[width=1.4cm]{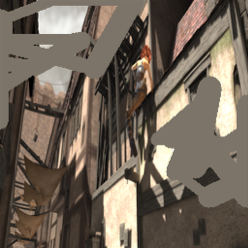}&
\includegraphics[width=1.4cm]{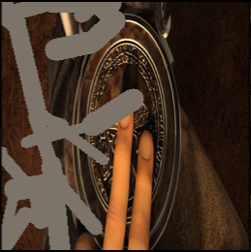}&
\includegraphics[width=1.4cm]{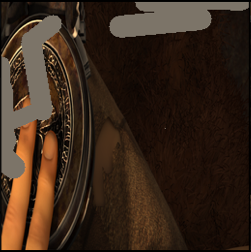}
\\

\addlinespace[-0.25em]

\rotatebox[origin=l]{90}{\scriptsize DeepFillv2}&
\includegraphics[width=1.4cm]{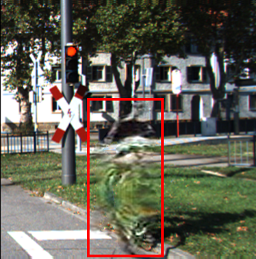}&
\includegraphics[width=1.4cm]{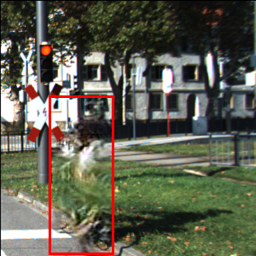}&
\includegraphics[width=1.4cm]{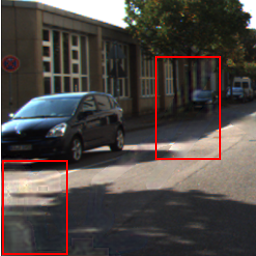}&
\includegraphics[width=1.4cm]{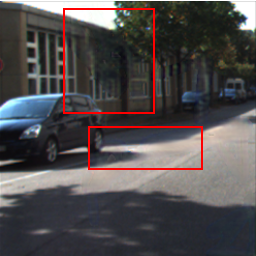}&
\includegraphics[width=1.4cm]{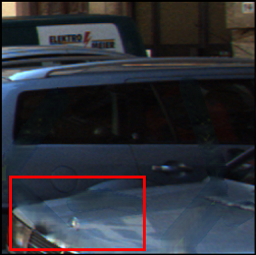}&
\includegraphics[width=1.4cm]{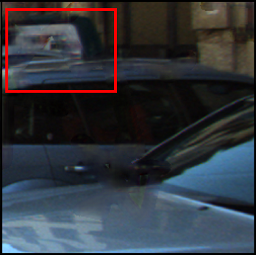}&
\includegraphics[width=1.4cm]{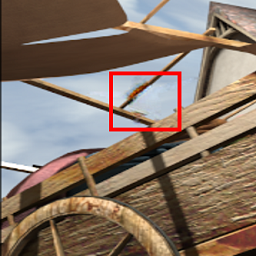}&
\includegraphics[width=1.4cm]{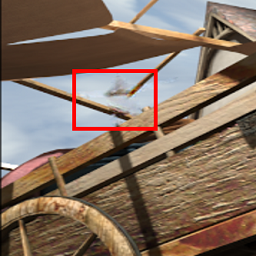}&
\includegraphics[width=1.4cm]{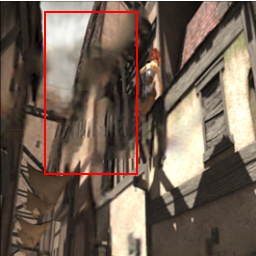}&
\includegraphics[width=1.4cm]{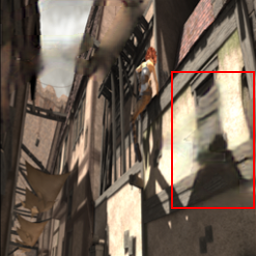}&
\includegraphics[width=1.4cm]{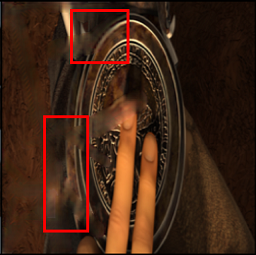}&
\includegraphics[width=1.4cm]{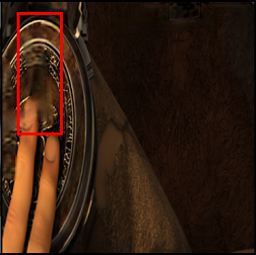}
\\

\addlinespace[-0.25em]

\rotatebox[origin=l]{90}{\scriptsize CR-Fill}&
\includegraphics[width=1.4cm]{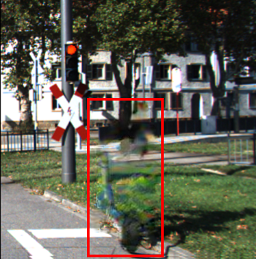}&
\includegraphics[width=1.4cm]{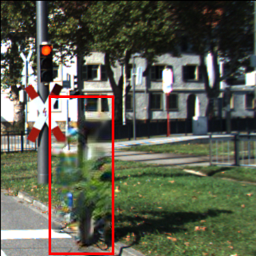}&
\includegraphics[width=1.4cm]{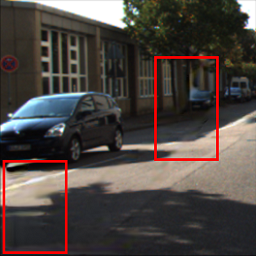}&
\includegraphics[width=1.4cm]{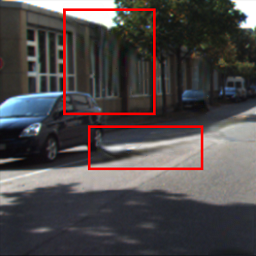}&
\includegraphics[width=1.4cm]{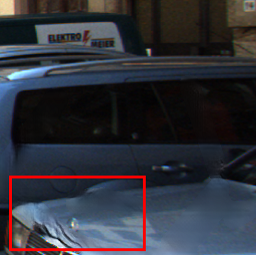}&
\includegraphics[width=1.4cm]{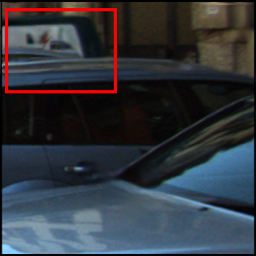}&
\includegraphics[width=1.4cm]{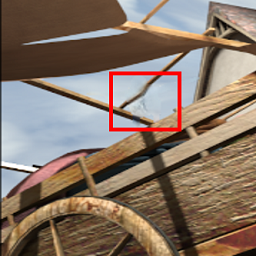}&
\includegraphics[width=1.4cm]{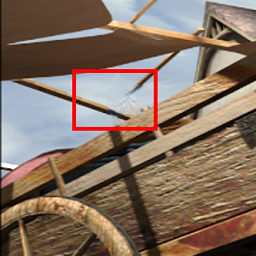}&
\includegraphics[width=1.4cm]{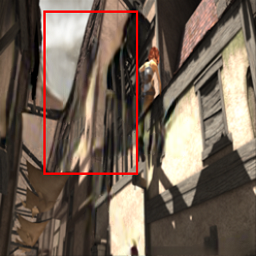}&
\includegraphics[width=1.4cm]{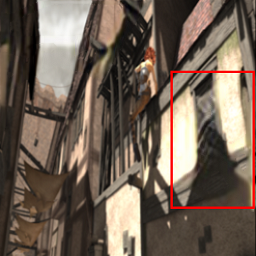}&
\includegraphics[width=1.4cm]{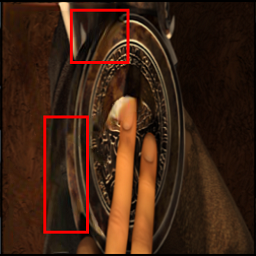}&
\includegraphics[width=1.4cm]{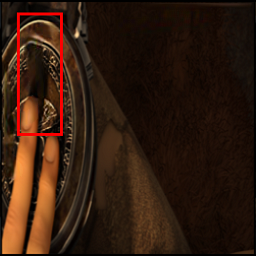}
\\

\addlinespace[-0.25em]

\rotatebox[origin=l]{90}{\scriptsize SICNet}&
\includegraphics[width=1.4cm]{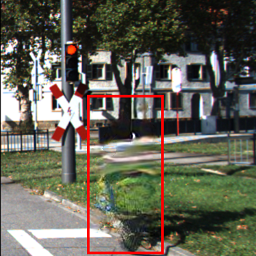}&
\includegraphics[width=1.4cm]{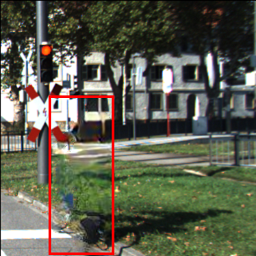}&
\includegraphics[width=1.4cm]{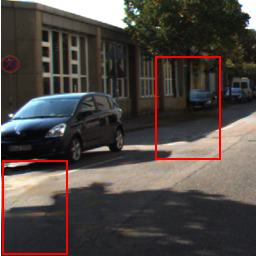}&
\includegraphics[width=1.4cm]{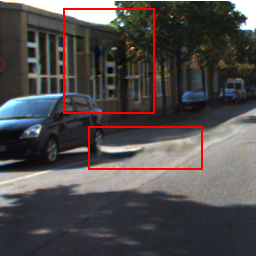}&
\includegraphics[width=1.4cm]{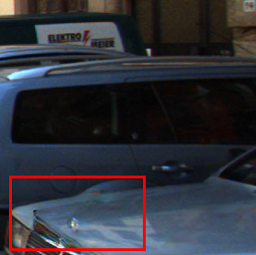}&
\includegraphics[width=1.4cm]{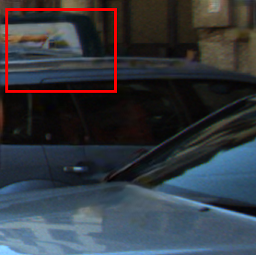}&
\includegraphics[width=1.4cm]{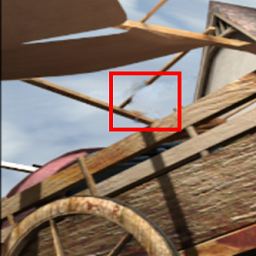}&
\includegraphics[width=1.4cm]{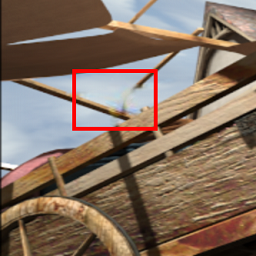}&
\includegraphics[width=1.4cm]{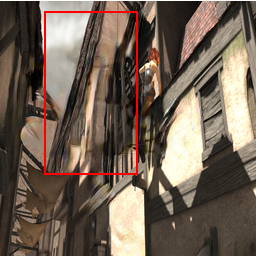}&
\includegraphics[width=1.4cm]{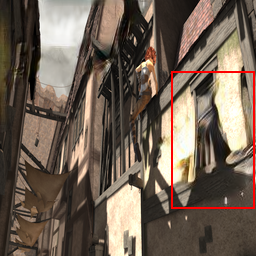}&
\includegraphics[width=1.4cm]{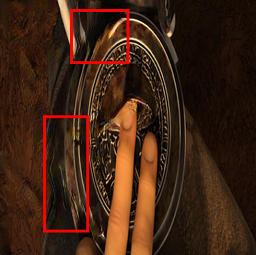}&
\includegraphics[width=1.4cm]{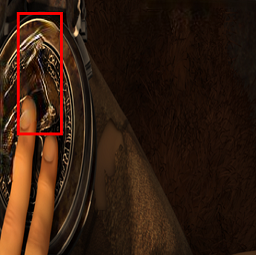}
\\

\addlinespace[-0.25em]

\rotatebox[origin=l]{90}{\scriptsize Ours}&
\includegraphics[width=1.4cm]{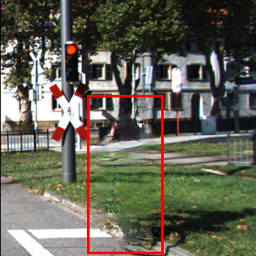}&
\includegraphics[width=1.4cm]{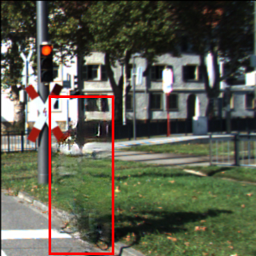}&
\includegraphics[width=1.4cm]{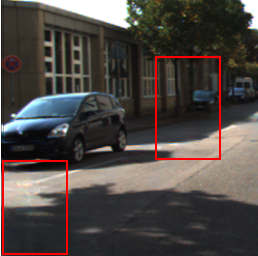}&
\includegraphics[width=1.4cm]{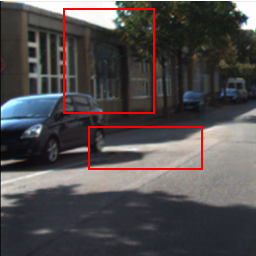}&
\includegraphics[width=1.4cm]{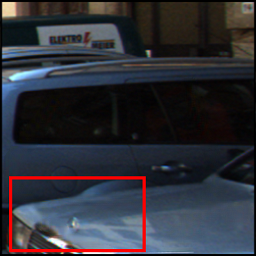}&
\includegraphics[width=1.4cm]{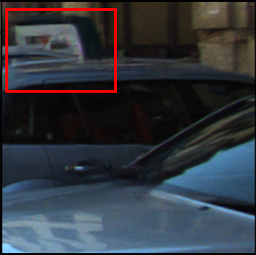}&
\includegraphics[width=1.4cm]{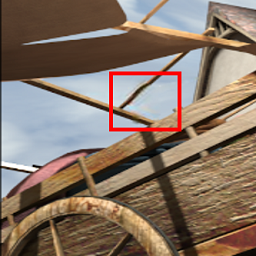}&
\includegraphics[width=1.4cm]{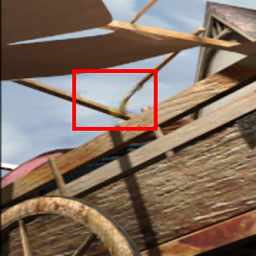}&
\includegraphics[width=1.4cm]{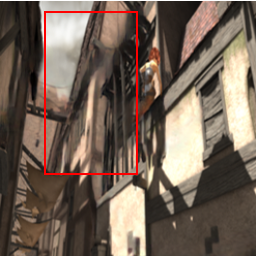}&
\includegraphics[width=1.4cm]{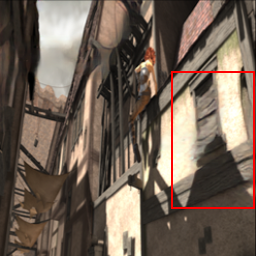}&
\includegraphics[width=1.4cm]{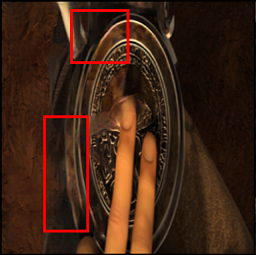}&
\includegraphics[width=1.4cm]{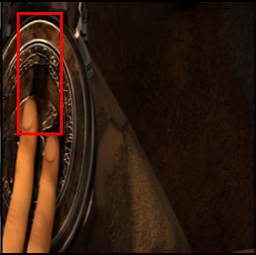}
\\

  &  \small Left   & \small Right   & \small Left   & \small Right & \small Left   & \small Right   & \small Left   & \small Right  & \small Left   & \small Right   & \small Left   & \small  Right

\end{tabular}  

}

\caption{Qualitative comparison of our proposed model with baseline models on KITTI2015 and MPI-Sintel datasets. Better viewed at zoom level 400\%. }
\label{QualitativeCompare}

\end{figure*}

To further evaluate the visual quality of the inpainted stereo images, we show three examples from the KITTI2015 dataset and three examples from the MPI-Sintel dataset in Figure~\ref{QualitativeCompare}.
Object-removal masks or irregular masks are applied on the ground truth images to obtain the corresponding input images.
It is obvious that both DeepFillv2 and CR-Fill struggle to maintain the stereo consistency of the inpainted views, which is a common problem for single-image inpainting models.
SICNet utilizes feature-fusion strategy to incorporate the geometry correlation during inpainting.
However, the fusion module directly concatenates the two views' feature maps,  which fails to effectively explore the stereo cues.
In contrast, our proposed IGGNet can generate not only vivid textures but also stereo consistent contents.

\subsection{Ablation Study}
We investigate the effectiveness of Geometry-Aware Attention (GAA) and Iterative Cross Guidance (ICG), respectively.

\begin{figure}[t!]
    \centering
    \setlength{\tabcolsep}{0.05em}
    {
\begin{tabular}{cccccc}
        
\rotatebox[origin=l]{90}{\small Left}&
\includegraphics[width=1.4cm]{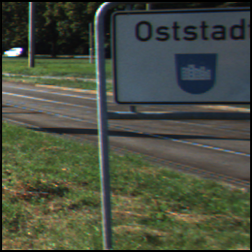}&
\includegraphics[width=1.4cm]{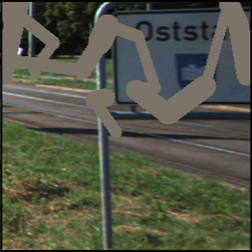}&
\includegraphics[width=1.4cm]{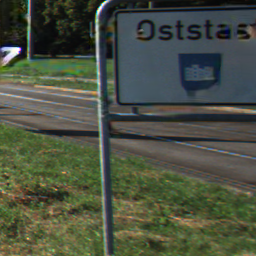}&
\includegraphics[width=1.4cm]{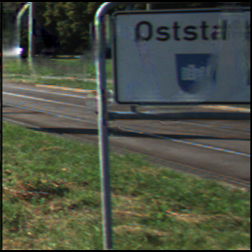}&
\includegraphics[width=1.4cm]{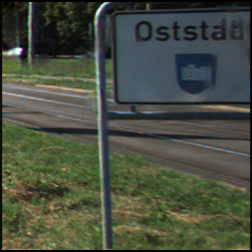}
\\

\addlinespace[-0.20em]

\rotatebox[origin=l]{90}{\small Right}&
\includegraphics[width=1.4cm]{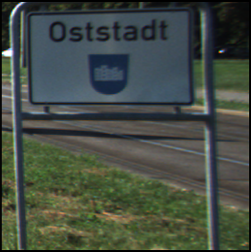}&
\includegraphics[width=1.4cm]{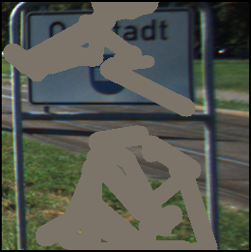}&
\includegraphics[width=1.4cm]{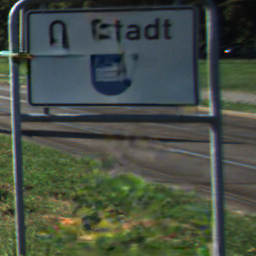}&
\includegraphics[width=1.4cm]{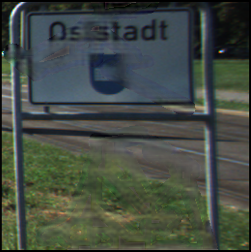}&
\includegraphics[width=1.4cm]{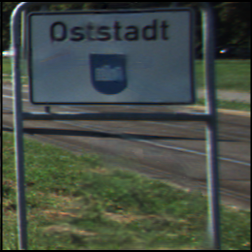}
\\

  &  \small GT & \small Input   & \small concat    & \small max    & \small  GAA  \\
   
\end{tabular}  

}
\caption{Ablation study on the GAA module. Better viewed at zoom level 400\%.  }
\label{ablation1}
\end{figure}

\subsubsection{Effectiveness of GAA}
As described in Sec.~\ref{secGAA}, the purpose of GAA is to effectively explore and integrate the geometry correspondence from $\phi_{ref}$ to guide the inpainting of $\phi_{tar}$.
We study the performance of our full model with two variants of GAA: 1) ``\textbf{concat}'', which directly concatenates $\phi_{ref}$ and $\phi_{tar}$ without constructing the 4D cost volume and attention-driven aggregation.
2) ``\textbf{max}'', which performs the 4D cost volume regression by directly choosing the maximum value on the disparity dimension and takes it as the input for the decoding network.
The performance drop of these two variants as shown 
in Table~\ref{Tab_ablation1} justifies the effectiveness of the GAA module.
As further shown in Figure~\ref{ablation1}, the model with GAA (concat) suffers from inaccurate feature alignment due to ignoring geometry correlation, thus produces distorted contents.
GAA (max) apparently improves the results, but it still brings negative impacts due to the simple feature aggregation.
In contrast, using full GAA leads to better results with geometry consistent details.

\begin{table}[thb]
\renewcommand\arraystretch{1.0}
\begin{center}
\scalebox{0.8}
{
\begin{tabular}{c|cccc}
      			\hline
      
    	& PSNR & SSIM  & FID  \\ \hline
   	concat	 & 25.17 & 0.863 & 10.73 \\ \hline
	max	   & 25.68 & 0.875  & 9.04 \\	 \hline
	GAA	   & \textbf{26.42} & \textbf{0.897} & \textbf{7.86} \\ \hline
		
\end{tabular}    
}
\end{center}
\caption{Ablation study on the GAA module over the KITTI2015 dataset with mask ratio $0-60\%$ and iteration number $T = 6$.}
\label{Tab_ablation1}
\end{table}

\subsubsection{Effectiveness of ICG}
\begin{figure}[t!]
    \centering
    \setlength{\tabcolsep}{0.05em}
    {
\begin{tabular}{cccccc}
        
\rotatebox[origin=l]{90}{\small Left}&
\includegraphics[width=1.4cm]{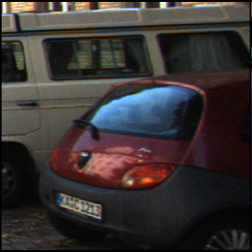}&
\includegraphics[width=1.4cm]{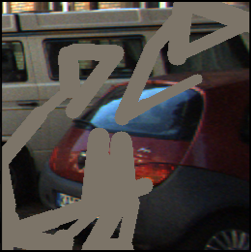}&
\includegraphics[width=1.4cm]{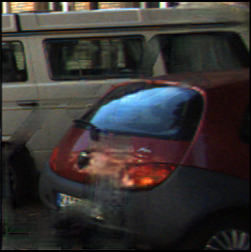}&
\includegraphics[width=1.4cm]{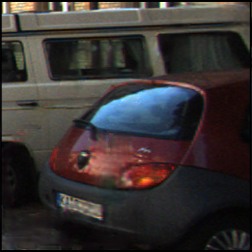}&
\includegraphics[width=1.4cm]{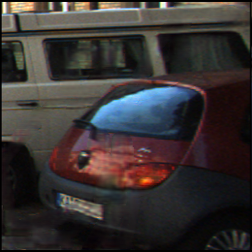}
\\

\addlinespace[-0.20em]

\rotatebox[origin=l]{90}{\small Right}&
\includegraphics[width=1.4cm]{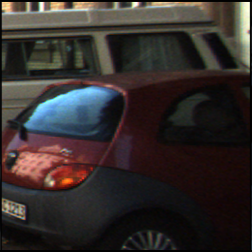}&
\includegraphics[width=1.4cm]{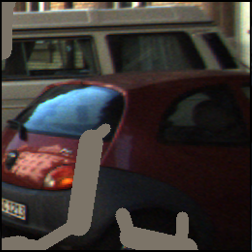}&
\includegraphics[width=1.4cm]{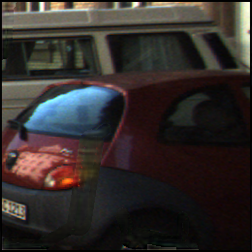}&
\includegraphics[width=1.4cm]{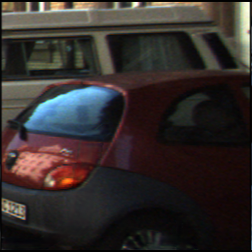}&
\includegraphics[width=1.4cm]{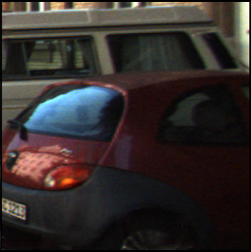}
\\

  &  \small GT & \small Input   & \small T = 4    & \small T = 6    & \small T = 10 \\
  
\end{tabular}  

}
\caption{Ablation study on different number of iterations $T$ of the ICG strategy. Better viewed at zoom level 400\%.  }
\label{ablation2}
\end{figure}

The ICG strategy aims to alternately use one view as the  reference to guide the inpainting of the other view in an iterative manner.
We inspect the impact of the number of iterations $T$.
Specifically, we trained our model with $T = 4, 6, 8, 10$. 
The evaluation results over the KITTI2015 dataset are shown in Table~\ref{Tab_ablation2}.
The best results are obtained when $T = 6$. 
A small $T$ (i.e., a small number of iterations) is insufficient to alleviate the co-existing missing region problem, leading to model performance degradation.
In contrast, a large $T$  shows no further performance improvement with the increase in computation time, because it  increases the depth and complexity of the whole procedure and makes model convergence more difficult and even overfitting.
An example is further shown in Figure~\ref{ablation2}.
When $T = 4$, the inpainting result suffers from visual artifacts and blurriness.
When $T = 6$ and $T = 10$, our model yields visual plausible results. 

\begin{table}[thb]
\renewcommand\arraystretch{1.0}
\begin{center}
\scalebox{0.8}
{
\begin{tabular}{c|ccccc}
      			\hline
      
    	&  T = 4 & T = 6 & T = 8 & T = 10    \\ \hline
   	PSNR	 &  25.98 & \textbf{26.42}   &  26.35   & 26.17 \\ \hline
	SSIM	   &  0.865 &  \textbf{0.879}  &  0.869  &  0.876 \\ \hline
		FID	   &  8.40 &  \textbf{7.86}  &  7.93  &  8.22 \\ \hline

\end{tabular}    
}
\end{center}
\caption{Ablation study on the $T$ value of ICG over the KITTI2015 dataset with mask ratio $0-60\%$.}
\label{Tab_ablation2}
\end{table}

\section{Conclusion}
In this work, we studied stereo image inpainting, attempting at inpainting the missing regions of the left and right image and maintaining their stereo consistency.
To achieve this, we first proposed a GAA module, which aims to associate the two images' features through learning an attention map by considering the epipolar geometry. Additionally, to address the co-existing missing regions, we further proposed an ICG strategy, which alternately inpaints one of two views. The whole network is named IGGNet. Taking advantage of the proposed modules, IGGNet outperforms the latest stereo image inpainting method and state-of-the-art single image inpainting methods.


\bibliographystyle{named}
\bibliography{ijcai22}

\end{document}